\documentclass[sigconf, nonacm]{acmart}
\AtBeginDocument{%
  }

\usepackage{xcolor}
\usepackage{multirow}
\usepackage{float}
\setcopyright{acmlicensed}
\copyrightyear{2018}
\acmYear{2018}
\acmDOI{XXXXXXX.XXXXXXX}

\acmISBN{978-1-4503-XXXX-X/2018/06}

\acmSubmissionID{1183}

\renewcommand\footnotetextcopyrightpermission[1]{}
\begin{document}

\title{Uni-HOI:  A Unified framework for Learning the Joint distribution of Text and Human-Object Interaction}


\author{Mengfei Zhang}
\authornote{Both authors contributed equally to this research.}
\email{zhangmengfei@whu.edu.cn}
\orcid{0009-0009-6736-4443}
\affiliation{%
  \institution{Wuhan University}
  \city{Wuhan}
  \country{China}
}


\author{Jinlu Zhang}
\authornotemark[1]
\email{jinluzhang@stu.pku.edu.cn}
\affiliation{%
  \institution{Peking University}
  \city{Beijing}
  \country{China}
}

\author{Zhigang Tu}
\authornote{Corresponding author.}
\email{tuzhigang@whu.edu.cn}
\affiliation{%
  \institution{Wuhan University}
  \city{Wuhan}
  \country{China}
}









\begin{abstract}
Modeling 4D human-object interaction (HOI) is a compelling challenge in computer vision and an essential technology powering virtual and mixed-reality applications.
While existing works have achieved promising results on specific HOI tasks—such as text-conditioned HOI generation and human motion generation from object motion, they typically rely on task-specific architectures and lack a unified framework capable of handling diverse conditional inputs. 
Building on this, we propose Uni-HOI, a unified framework that learns the joint distribution among text, human motion, and object motion. By leveraging large language models (LLMs) and two motion-specific vector quantized variational autoencoders (VQ-VAEs), we convert heterogeneous motion data into token sequences compatible with LLM inputs, enabling seamless integration and joint modeling of all three modalities. We introduce a two-stage training strategy: the first stage performs multi-task learning on a large-scale HOI dataset to capture the underlying correlations among the three modalities, while the second stage fine-tunes the model on specific tasks to further enhance performance.
Extensive experiments demonstrate that Uni-HOI achieves remarkable performances on multiple HOI-related tasks including text-driven HOI generation, object motion-driven human motion generation (optionally with text) and human motion-driven object motion prediction within a unified framework.
\end{abstract}




\begin{teaserfigure}
  \includegraphics[width=\textwidth]{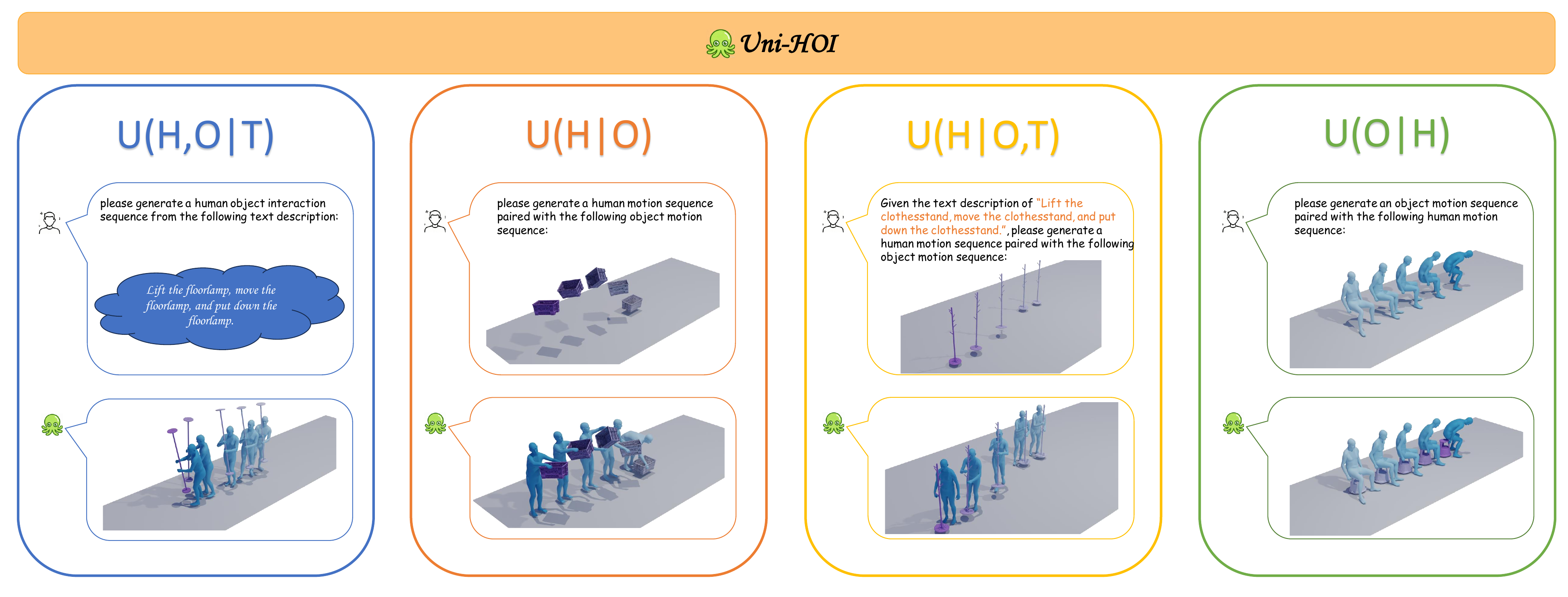}
  \caption{Uni-HOI learns the joint distribution of text and human-object interaction, thus it can serve multiple HOI-related tasks including text-driven interaction generation, object-guided motion generation(optionally with text) and object motion prediction. \textbf{T}, \textbf{H} and \textbf{O} represent the text, human motion and object motion.}
  \label{fig:teaser}
\end{teaserfigure}


\maketitle

\section{Introduction}
Every day, humans interact with a wide variety of objects to perform daily tasks. Accurately understanding and modeling these interactions, and furthermore, generating the motions of both the human and the objects involved, is of great significance to fields such as virtual reality~\cite{lee2026hoicraft}, embodied artificial intelligence~\cite{jia2025primhoi,jung2025dechoi}, gaming~\cite{hu2022hogan}, and animation~\cite{cao2024avatargo}.

In the existing field of Human-Object Interaction (HOI), numerous studies have explored a variety of tasks. For instance, works such as HOI-Diff~\cite{hoi-diff} and ROG~\cite{rog} have investigated text-conditioned HOI generation, OMOMO~\cite{omomo} and SemGeoMo~\cite{semgeomo} explore the generation of human motion based on object motion, and Interdiff~\cite{xu2023interdiff} addresses the task of HOI prediction. While these methods have achieved promising results on specific tasks within the HOI domain, there remains a lack of a unified framework capable of properly modeling the joint distribution among text, human motion, and object motion—thereby enabling a single framework to accomplish multiple tasks in the HOI field. TriDi~\cite{tridi} attempts to address this by employing a diffusion model to generate human, object, and interaction modalities simultaneously through a novel three-way diffusion process. However, its application is limited to 3D HOI generation and does not extend to the broader scope of 4D HOI scenarios. By simultaneously modeling all modalities, such a joint framework can process arbitrary conditional inputs based on specific requirements, thereby accomplishing multiple HOI tasks with a unified framework. This a capability that previous works, relying on a single model, could not achieve.
By jointly modeling text, human motion, and object motion within a single architecture, it eliminates the need for task-specific model designs, thereby reducing training complexity and improving parameter efficiency. This unified framework inherently supports flexible conditional generation—given any subset of modalities as input, it can coherently infer the missing ones, enabling seamless adaptation to diverse HOI task.

Building on this, we propose Uni-HOI, a unified framework designed to learn the joint distribution among text, human motion, and object motion, thereby serving a variety of tasks within the HOI domain. With the rapid advancement of large language models (LLMs) and the success of Motion-GPT~\cite{motiongpt}, leveraging LLMs to construct such a unified framework presents a promising direction. 
To this end, we first learn two motion-specific vector quantized variational autoencoder (VQ-VAE)~\cite{vqvae} models to construct "human motion vocabulary" and "object motion vocabulary". These two models can convert human motion and object motion into special sequence tokens. These tokens share a representational format analogous to the word tokens of LLMs, enabling them to be integrated and fed into a large language model for fine-tuning. Subsequently, these tokens are fed into a pre-trained large language model~\cite{chung2024scaling,embedding-exploring,yang2024fhoi}, which captures the underlying grammar and syntax of the human-object interaction language and learns how it relates to the corresponding textual descriptions. 

Uni-HOI is first trained on the largest HOI dataset collected to date by randomly combining certain modalities as conditions to generate the remaining ones. In this process, leveraging the powerful modeling capabilities of large language models, Uni-HOI learns the associations among three distinct data distributions of human motion, object motion, and text. To further enhance performance of the model on various specific HOI tasks, we conduct a second-stage training for Uni-HOI under the input-output configurations of each specific task. 
We demonstrate the superiority of Uni-HOI, which establishes the first unified framework for learning and aligning the data distributions of human motion, object motion, and text in the HOI domain. Furthermore, Uni-HOI is capable of handling flexible inputs and accordingly generating the desired human motion, object motion or text. Remarkably, with only a single framework, Uni-HOI achieves superior performance over specially designed baselines across multiple HOI-related tasks.

We summarize our contributions as follows: 

(1)We propose Uni-HOI, the first unified framework in the HOI domain capable of aligning the data distributions among text, human motion, and object motion. It can handle arbitrary combinations of these three modalities as input and generate corresponding outputs based on task requirements, thereby serving a variety of HOI tasks. 

(2) We propose a two-stage training strategy, which first performs multi-task training to learn the data distribution relationships between text and HOI, followed by a second stage of fine-tuning on specific tasks to further enhance the model's performance on those particular tasks.

(3) Extensive experimental results demonstrate that Uni-HOI, with only a single framework, outperforms specially designed baselines across multiple HOI tasks and highlights its superior performance and the significance of our proposed approach.


\begin{figure*}[htbp]
    \centering
    \includegraphics[width=\textwidth]{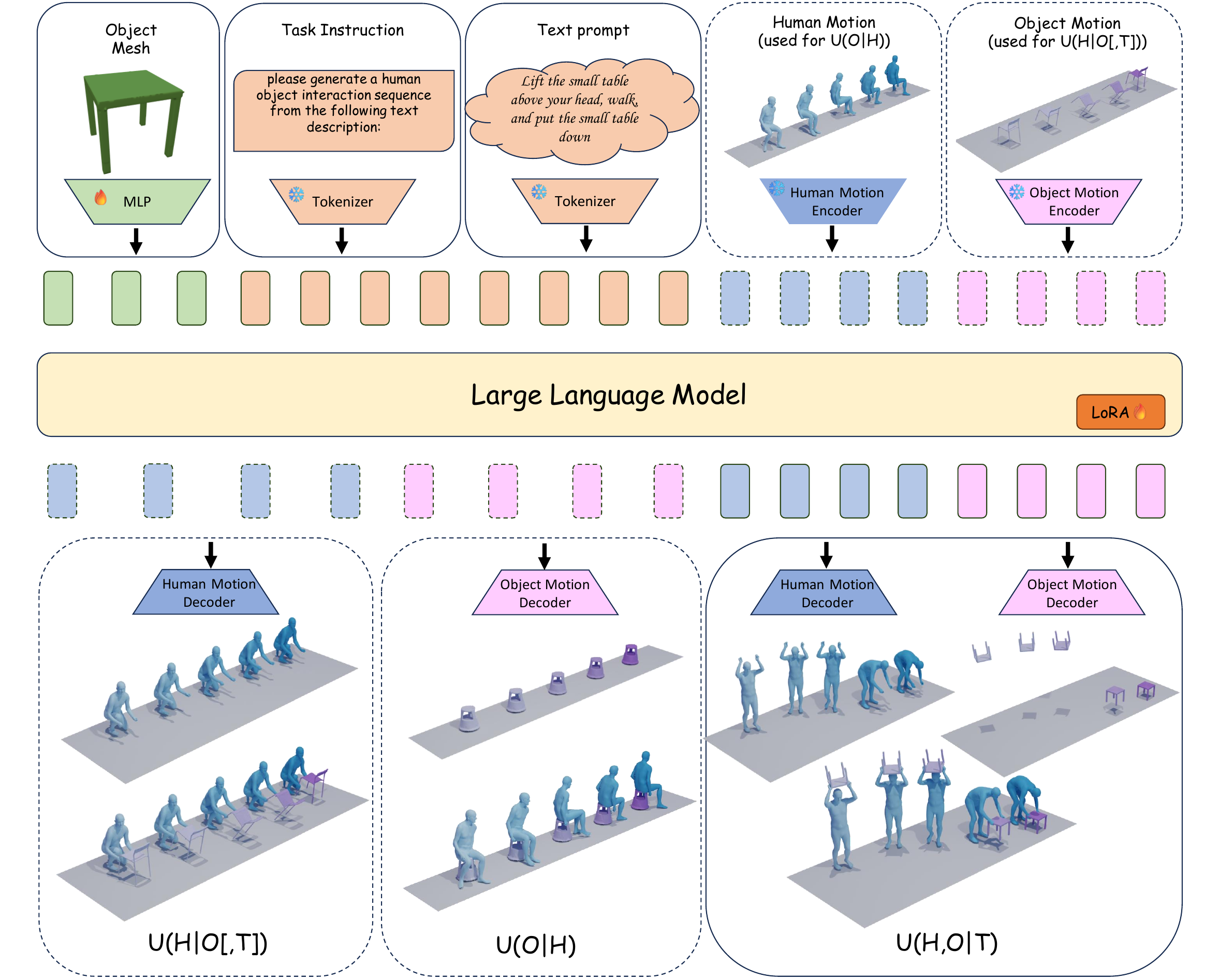}
    \caption{Overview of our Uni-HOI framework. Based on different task specifications, Uni-HOI can serve a variety of HOI tasks, including text-driven HOI generation (U(H,O|T)), human motion generation guided by object motion(optionally with text)(U(H|O[,T])) and object motion prediction guided by human motion(U(O|H)).}
    \label{fig:pipline}
\end{figure*}


\section{Related Work}

\textbf{Text-driven Human-Object Interaction Generation}.  Within the HOI domain, a particularly prominent and widely studied area is text-driven HOI generation. This task aims to synthesize semantically coherent and physically plausible human-object interaction sequences using only text as input, holding significant potential for applications in gaming and animation~\cite{hoi-diff,dai2024interfusion,wu2024thor, cai2025interactmove,xu2024interdreamer,li2025genhoi,wu2025hoifhi,li2024chois,geng2025auto}. Numerous prior works have explored this direction. For example, HOI-Diff~\cite{hoi-diff} and OOD-HOI~\cite{ood-hoi} investigate the use of diffusion models followed by post-processing optimization to generate HOI from text. CG-HOI~\cite{cg-hoi} synthesizes interactions using a single diffusion model with explicit contact modeling. 
Text2HOI~\cite{cha2024text2hoi} introduces the first text-guided work for generating the sequence of hand-object interaction in 3D. 
OOD-HOI~\cite{ood-hoi}, in particular, focuses on generating human hand manipulations of diverse objects, with an emphasis on hand-object contact.
ROG~\cite{rog} introduces a diffusion-based framework to synthesize realistic human-object interactions by modeling spatiotemporal relationships with rich geometric details from text description. 
It proposes an Interactive Distance Field (IDF) constructed from boundary-focused and finely sampled object key points to capture comprehensive spatial dynamics. 
These specially designed methods can only take text as conditioning and are unable to synthesize HOI under more general conditional settings.

\noindent \textbf{Object motion-driven Human Motion Generation}. In the field of Human-Object Interaction (HOI), another research task that has garnered widespread attention is the generation of human motion guided by object motion. OMOMO~\cite{omomo} achieved this through a two-stage approach that first predicts hand positions from object geometry and then generates full-body poses conditioned on the corrected hand positions with explicit contact constraints. 
BimArt~\cite{bimart} introduces a unified generative approach for synthesizing realistic 3D bimanual hand interactions with articulated objects from the object’s 7 DoF trajectory (6D global states + 1D articulation), eliminating the need for pre-defined grasps, coarse hand trajectories, or category-specific training. 
SemGeoMo~\cite{semgeomo} proposes a novel approach for dynamic contextual human motion generation that generates semantically rational and geometrically accurate human interactive motions alongside corresponding textual descriptions from 4D sequential point clouds of dynamic interactive targets, addressing the limitations of prior work in lacking fine-grained semantic and geometric guidance. Uni-HOI utilizes a single unified network to not only generate human motion from object motion, but also to perform object motion generation from human motion and text.

\noindent \textbf{Joint modeling Human-Object Interaction}. TriDi~\cite{tridi} presents the first unified probabilistic model for 3D human-object interaction (HOI) that jointly models the three modalities of Human (H), Object (O), and Interaction (I), enabling sampling from seven conditional and joint distributions (e.g., P(H,I|O), P(O,I|H), P(H,O,I)) with a single network, thus unifying and extending prior one-way specialized methods. However, TriDi is limited to modeling 3D HOIs, a task that is considerably less challenging and has a narrower range of applications compared to 4D HOI tasks involving sequences. Furthermore, TriDi utilizes contact maps instead of more user-friendly text for modeling, which significantly diminishes its practical value. Uni-HOI is the first to simultaneously model text, human motion, and object motion in the 4D HOI domain, enabling a single model to serve multiple 4D HOI tasks.

\section{Method}

\textbf{Overview}. Our objective is to leverage the powerful modeling capabilities of large language models to construct a unified framework for the HOI domain, aimed at learning the distributional relationships among text, human motion, and object motion and serving for multiple HOI tasks. To enable the integration of human motion and object motion with text for joint modeling within a large language model, we first train two VQVAE~\cite{vqvae} models, serving as a human motion tokenizer and an object motion tokenizer, respectively. These two VQVAEs convert human motion and object motion into discrete motion tokens, which can act as two distinct "foreign languages" alongside text tokens, allowing a pre-trained large language model~\cite{embedding-exploring,embedding-sentencepiece,embedding-training,qwen3,llama,chatgpt} to learn the correspondences among the three modalities. To flexibly accommodate various input-output configurations and enhance performance on specific tasks, we propose a two-stage training strategy. Uni-HOI is first trained on the largest collected HOI dataset, tailored to the settings of multiple mainstream HOI tasks, aiming to learn the relationships among the three modalities under different input-output configurations. To further improve its performance on specific HOI task, we introduce a second-stage fine-tuning strategy, conducting additional training based on the task-specific input-output configurations of the three modalities to boost performance on those specific tasks.

\subsection{Preliminary on Human-Object Interaction}
Human-object interaction primarily consists of a human motion sequence and an object motion sequence, aiming to model the interaction process between humans and various objects in the real world. The human motion sequence is represented by a series of human pose parameters, denoted as \(H\) = \( \{h_i\} \in \mathbb{R}^{L \times D_h} \),  where \( L \) is the length of the sequence and \( D_h \) is a parameterized representation of human pose using the SMPL-H~\cite{smplh} model with \( D_h = 159 \), which includes root translation, orientation, and rotations of other joints in the pelvis space. For the object motion, it is obtained from the object mesh in a static state through a series of translation and rotation transformations. The object motion is denoted as \(O\) = \( \{o_i\} \in \mathbb{R}^{L \times D_o} \), in which \( D_o = 6 \). The object mesh is given as condition and we sample each object mesh to a point cloud consisting of 340 points, which is denoted as \( P \in \mathbb{R}^{340 \times 3} \).

\subsection{Motion Tokenizer}

In order to integrate the features of human motion and object motion into a large language model, we pre-train two motion tokenizers \(V_h\) and \(V_o\) based on the Vector Quantized Variational Autoencoder (VQ-VAE)~\cite{vqvae} architecture adopted in~\cite{t2mgpt,tm2t,motiongpt,bailando,van2017neural}. Our motion tokenizer comprises an encoder \(E\) and a decoder \(D\). The encoder generates discrete motion tokens with high information density, while the decoder reconstructs these tokens back into motion sequences \(\hat{\mathbf{M}}_{1:L}\)  . This paradigm enables efficient representation of motion as a discrete language, thereby facilitating the integration of motion and language for a variety of motion-related tasks.

Specifically, the motion encoder \(E\) first applies 1D convolutions along the temporal dimension to the frame-wise motion features, producing latent vectors \(\hat{\mathbf{z}}_{1:m} = E(\mathbf{M}_{1:L})\). Subsequently, \(\hat{\mathbf{z}}\) is transformed into a collection of codebook entries \(\mathbf{z}\) via discrete quantization. The learnable codebook \(\mathcal{Z} = \{\mathbf{z}_k\}_{k=1}^{K} \subset \mathbb{R}^d\) consists of \(K\) latent embedding vectors, each of dimension \(d\). The quantization operation \(Q(\cdot)\) replaces each latent vector \(\hat{\mathbf{z}}_i\) with its nearest codebook entry \(\mathbf{z}_k\) in \(\mathcal{Z}\), formalized as:

\begin{equation}
\mathbf{z}_i = Q(\hat{\mathbf{z}}_i) := \arg \min_{\mathbf{z}_k \in \mathcal{Z}} \| \hat{\mathbf{z}}_i - \mathbf{z}_k \|_2
\end{equation}

After quantization, the motion decoder \(D\) projects the quantized sequence \(\mathbf{z}_{1:m} = \{\mathbf{z}_i\}_{i=1}^{m}\) back to the original motion space, reconstructing the motion sequence \(\hat{\mathbf{M}}_{1:L}\) with \(L\) frames.

To train this motion tokenizer, we follow~\cite{t2mgpt,motionmillion} in employing four distinct loss functions:

\begin{equation}
\mathcal{L}_V = \mathcal{L}_r + \mathcal{L}_e + \mathcal{L}_c + \mathcal{L}_v
\end{equation}

\noindent where \(\mathcal{L}_r\) denotes the reconstruction loss, \(\mathcal{L}_e\) the embedding loss, and \(\mathcal{L}_c\) the commitment loss. To further enhance the quality of generated motion, we incorporate velocity regularization \(\mathcal{L}_v\) into the reconstruction loss, as suggested by~\cite{t2mgpt}. Additionally, we adopt exponential moving average (EMA) and codebook reset techniques~\cite{vqvae2} to improve codebook utilization during training. In particular, our human motion tokenizer \(V_h\) and object motion tokenizer \(V_o\) share the same VQ-VAE model architecture but differ in their input dimensions and the feature dimensions of their encoders.

\subsection{Uni-HOI: Unified framework for Text-Human-Object Modeling}
\textbf{Joint modeling}. Our goal is to use a unified framework for the joint modeling of text, human motion, and object motion, thereby enabling flexible processing of multiple inputs and generating desired outputs according to the requirements of various downstream tasks for HOI.

\begin{equation}
\mathcal{T} = \left\{ (\mathcal{X}, \mathcal{Y}) \;|\; \mathcal{X} \subset \mathcal{Q}, \mathcal{Y} = \mathcal{Q} \setminus \mathcal{X}, \mathcal{X} \notin \{\emptyset, \mathcal{Q}\} \right\}
\end{equation}

\begin{equation}
\theta^* = \arg\min_{\theta} \sum_{(\mathcal{X}, \mathcal{Y}) \in \mathcal{T}} \mathbb{E}_{(P, T, H, O) \sim \mathcal{D}} \left[ \ell\left( \mathcal{Y}, U_\theta(P, \mathcal{X}) \right) \right]
\end{equation}

\noindent where \(\mathcal{Q} = \{T, H, O\}\) and \(T \) represents the text condition, 
\(H\) is the human motion, \(O\) is the object motion. \(\mathcal{X}\) denotes one or two modalities from \(\mathcal{Q}\) as conditions, and \(\mathcal{Y}\) denotes the remaining modalities. \(\ell\) is the loss function and \(P\) represents the sampled object points.

Employing the two motion tokenizers, a human motion sequence \(\mathbf{H}_{1:L}\) and an object motion sequence \(\mathbf{O}_{1:L}\) can be mapped to a sequence of human motion tokens \(\mathbf{z}_{1:m}^h\) and object motion tokens \(\mathbf{z}_{1:m}^o\), allowing joint representation with vocabulary embeddings in language models~\cite{embedding-exploring,embedding-sentencepiece,embedding-training}. By integrating them into a unified vocabulary, we learn motion and language jointly. We first represent human motion tokens \(\mathbf{z}_{1:m}^h\) and object motion tokens \(\mathbf{z}_{1:m}^o\) as a sequence of indices \(\mathbf{s}_{1:m}^h = \{s_i^h\}_{i=1}^{m}\) and \(\mathbf{s}_{1:m}^o = \{s_i^o\}_{i=1}^{m}\), where \(s_i^h\) and \(s_i^o\) correspond to the indexes of human motion token \(\mathbf{z}_i^h\) and object motion \(\mathbf{z}_i^o\) in the codebook of motion tokenizers.

Following MotionGpt~\cite{motiongpt}, we combine the original text vocabulary \(\mathcal{V}_t = \{v_i^t\}_{i=1}^{K_t}\) with a human motion vocabulary \(\mathcal{V}_h = \{v_i^h\}_{i=1}^{K_h}\) as well as an object motion vocabulary \(\mathcal{V}_o = \{v_i^o\}_{i=1}^{K_o}\), which preserves the order of our motion codebooks \(\mathcal{Z}_h\) and \(\mathcal{Z}_o\). Moreover, \(\mathcal{V}_h\) includes special tokens such as boundary indicators \(\texttt{<bohm>}\), \(\texttt{<eohm>}\), \(\texttt{<boom>}\) and \(\texttt{<eoom>}\) denoting the beginning and end of motion sequence. Thus, we construct a unified text-hoi vocabulary \(\mathcal{V} = \{\mathcal{V}_t, \mathcal{V}_h,\mathcal{V}_o \}\), enabling diverse motion-related tasks to be formulated in a general format where both input and output "words" are drawn from the same \(\mathcal{V}\). These "words" may represent natural language, human motion or object motion, depending on the task at hand. Consequently, our Uni-HOI supports flexible representation and generation of various text-HOI related outputs within a single framework.

To address conditioned generation tasks, we adopt a transformer-based architecture following Qwen3~\cite{qwen3}, which effectively maps input sequences to outputs. As illustrated in Figure \ref{fig:pipline}, our input source is a sequence of tokens \(\mathbf{X}_s = \{x_p,\{x_{s_i}\}_{i=1}^{N}\}\), where \(x_s \in \mathcal{V}\) and \(N\) denotes the input length. As for \(x_p\), we design an MLP network to extract features from the input object point cloud, which is then inserted into the input sequence as a special \textit{<object geometry token>} to serve as the embedding of geometric information of interacted object. And the target output is \(\mathbf{X}_t = \{x_{t_i}\}_{i=1}^{L}\), with \(x_t \in \mathcal{V}\) and \(L\) the output length. Then the source tokens are fed into the transformer encoder, and the decoder subsequently predicts the probability distribution of the next token at each step in an autoregressive manner:

\begin{equation}
p_\theta(\mathbf{X}_t | \mathbf{X}_s) = \prod_{i=0}^{L-1} p_\theta\left(x_{t_i} \mid x_{t}^{<i}, \mathbf{X}_s\right),
\end{equation}

where \(x_t^{<i}\) denotes the tokens preceding position \(i\). During training, we maximize the log-likelihood of the data distribution:

\begin{equation}
\mathcal{L}_{\mathrm{LM}} = -\sum_{i=0}^{L-1} \log p_\theta\left(x_{t_i} \mid x_t^{<i}, \mathbf{X}_s\right).
\end{equation}

By optimizing this objective, Uni-HOI learns to capture underlying patterns and dependencies from the data distribution of three modalities, facilitating accurate and meaningful generation of target "words". During inference, target tokens are recursively sampled from the predicted distribution \(p_\theta(\hat{x}_{t_i} \mid \hat{x}_t^{<i}, \mathbf{X}_s)\) until the end token is generated. This sampling strategy enables step-by-step generation, where each token is probabilistically determined based on previously generated tokens and the given source input.


\noindent \textit{Two-stage training strategy}. We first conduct a preliminary training stage for Uni-HOI on the largest collected HOI dataset. The objective is to model the joint distribution across text, human motion, and object motion by employing randomized combinations of the three modalities as input-output pairs. Specifically, given paired text-human motion-object motion data, any one or two of the modalities are designated as conditions, and the model is tasked with predicting the remaining modality based on the provided task instruction and conditions. For example, in the text-driven human-object interaction generation task, a task instruction could be: "\textit{Please generate a human-object interaction sequence from the following text description}". The model is then required to predict the corresponding indices of human motion and object motion within the two motion codebooks. Similarly, for object motion-guided human motion generation, an instruction prompt could be: "\textbf{Please generate a human motion sequence paired with the following object motion sequence:} \textit{<object motion tokens>}".
To further enhance the performance of Uni-HOI across various downstream tasks, we conduct adaptive second-stage training on each individual sub-task. Confined to a single HOI task, this second-stage training is more effective in improving its performance under fixed input-output conditions. We used the LoRA~\cite{hu2022lora} algorithm to fine-tune the backbone of the large language model in both training stages, and adopt a smaller rank and scaling coefficient during the second-stage training.


\begin{table*}[htbp]
    \centering
    \caption{Quantitative experimental results for the task of text-driven HOI generation on FullBodyManipulation and BEHAVE dataset. → means results closer to the real distribution are better.}
    \setlength{\arrayrulewidth}{0.1em} 
    \begin{tabular}{l|ccccc|ccccc}
        \hline
        & \multicolumn{5}{c|}{FullBodyManipulation} & \multicolumn{5}{c}{BEHAVE} \\
        \hline
        \multirow{2}{*}{Methods} 
        & \multirow{2}{*}{FID ↓} 
        & \multicolumn{3}{c}{R-precision ↑} 
        & \multirow{2}{*}{Diversity →} 
        & \multirow{2}{*}{FID ↓} 
        & \multicolumn{3}{c}{R-precision ↑} 
        & \multirow{2}{*}{Diversity →} \\
        \cline{3-5} \cline{8-10}
        & & Top-1 & Top-2 & Top-3 & & & Top-1 & Top-2 & Top-3 & \\
        \hline
        GT      & 0.01 & 0.53 & 0.63 & 0.69 & 7.82 & 0.09 & 0.75 & 0.82 & 0.89 & 7.76 \\
        MDM*~\cite{mdm}& 8.47 & 0.25 & 0.32 & 0.41 & 8.82 & 0.52 & 0.17 & 0.19 & 0.22 & 7.46 \\
        HOI-Diff~\cite{hoi-diff} & 11.25 & 0.21 & 0.27 & 0.31 & 10.23 & 0.59 & 0.16 & 0.20 & 0.23 & 7.14 \\
        ROG~\cite{rog}     & 5.35 & 0.42 & 0.48 & 0.52 & 8.52 & 0.42 & 0.30 & 0.37 & 0.41 & 7.55 \\
        Uni-HOI & \textbf{5.13} & \textbf{0.45} & \textbf{0.51} & \textbf{0.55} & \textbf{8.07} & \textbf{0.37} & \textbf{0.39} & \textbf{0.42} & \textbf{0.48} & \textbf{7.62} \\
        \hline
    \end{tabular}

    \label{tab:t2hoi}
\end{table*}

\begin{figure*}[htbp]
    \centering
    \includegraphics[width=\textwidth]{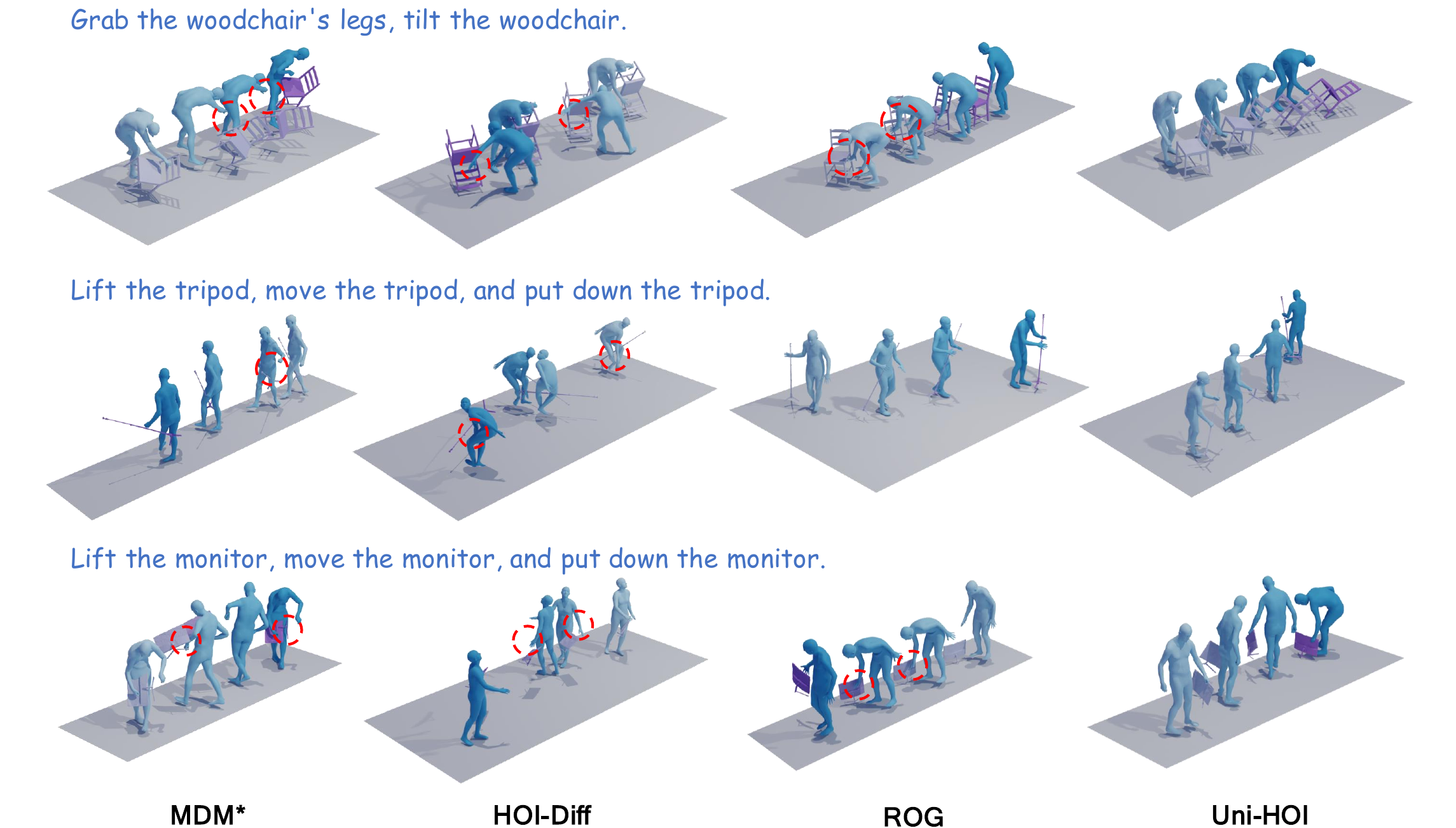}
    \caption{Visualization of comparative results for the task of text-driven HOI generation on the FullBodyManipulation and BEHAVE datasets. Irrational interactions are circled in red.}
    \label{fig:t2hoi}
\end{figure*}

\section{Experiments}

Extensive experimental results demonstrate the superior performance of Uni-HOI over baseline methods across multiple HOI-related tasks and datasets. In the following, we first introduce the datasets used in the experiments, evaluation metrics, and implementation details. Subsequently, we report both qualitative and quantitative experimental results comparing Uni-HOI with baselines on four mainstream HOI tasks, thereby providing a comprehensive analysis of its performance.

\subsection{Experimental Setup}

\noindent \textbf{Datasets}. Interact~\cite{interact} constructs a large-scale 4D human-object interaction dataset by integrating seven existing datasets (including BEHAVE~\cite{behave}, OMOMO~\cite{omomo}, GRAB~\cite{grab}, etc.) and it comprises 21.81 hours and 16,201 sequences in total. Due to the task setup and the quality of text annotations, we excluded data from ARCTIC~\cite{fan2023arctic} and Parahome~\cite{kim2025parahome}, and ultimately used a subset of InterAct that includes data from BEHAVE~\cite{behave}, Chairs~\cite{chairs}, GRAB~\cite{grab}, OMOMO~\cite{omomo}, InterCap~\cite{huang2024intercap}, IMHD~\cite{IMHD}, and NeuralDome~\cite{zhang2023neuraldome}. By partitioning each sub-dataset into training and test sets, the data used to train Uni-HOI comprises a total of 10,485 sequences for training and 1,274 sequences for testing.


\noindent \textbf{Evaluation Metrics.} (1) For the text-driven HOI generation, we evaluated using \textbf{Fréchet Inception Distance} (FID), \textbf{R-Precision} and \textbf{Diversity} as evaluation metrics. \textbf{FID} assesses the distribution difference between the generated HOI and the ground truth, while \textbf{R-Precision} reports the degree of alignment between the generated HOI and the text. \textbf{Diversity} captures the degree of variation present in the generated motions.
(2) For the object motion-guided human motion generation, our evaluation primarily follows the metrics established in OMOMO~\cite{omomo}. \textbf{HandJPE} and \textbf{MPJPE} denote the mean hand joint position error and the mean per‑joint position error, respectively, both computed as the Euclidean distance between the predicted and ground‑truth positions in centimeters (cm). To assess interaction quality, we adopt contact metrics—precision (\(C_{prec}\), recall (\(C_{rec}\)) and accuracy (\(C_{acc}\)) following~\cite{omomo}. The contact percentage (\(c\%\)) represents the proportion of frames in which contact is detected.
(3) For the human motion-guided object motion prediction, we follow \cite{objpop} to report the
\textbf{\(\mathbf{E_{v2v}}\)} and \textbf{\(\mathbf{E_{ch}}\)} as evaluation metrics. \textbf{\(\mathbf{E_{v2v}}\)} computes the error between predicted object points and the ground truth as a point-to-point error. \textbf{\(\mathbf{E_{ch}}\)} calculates the bi-directional Chamfer distance between the prediction and ground truth.


\noindent \textbf{Implementation Details.} For human VQ-VAE and object VQ-VAE, the codebook is set to be $K \in \mathbb{R}^{512 \times 4096}$. The motion encoder \(E\) incorporates a temporal downsampling rate l of 2. We utilize Qwen3-8B as the underlying architecture for our language model. 
The model consists of 36 transformer layers, with grouped query attention (GQA) where the number of query heads is 32 and the number of key/value heads is 8.
The feed-forward networks have an output dimensionality of \(d_{ff}\) =25600, and the attention mechanisms employ an inner dimensionality of \(d_{kv}\) =64. 
The remaining sub-layers and embeddings have a dimensionality of \(d_{model}\) =4096. 
Moreover, all our models employ the AdamW~\cite{adamw} optimizer for training. The human motion tokenizer and object motion tokenizer are trained with a learning rate of \(2e^{-4}\) for 1.5 million and 1 million iterations. Both tokenizers are trained with a batch size of 128. The language model is trained with LoRA~\cite{hu2022lora}, of which the rank and $ \alpha $ are set to be 1024 and 512 with a learning rate of \(2e^{-5}\) in the first stage. For the second-stage training, the rank and $ \alpha $ are set to be 256 and 128 with a learning rate of \(1e^{-6}\). The first stage of training has a total iteration of 110K and the second stage has 50K iterations. All models are trained on 8 NVIDIA GeForce RTX 4090 GPUs.


\subsection{Comparisons on HOI-relevant Tasks}
\noindent \textbf{Comparisons on Text-to-HOI.} In the second stage of training, we performed fine-tuning on the text-to-HOI task using the FullBodyManipulation~\cite{omomo} and BEHAVE~\cite{behave} datasets. We compared our method with several open-source baselines, including MDM*~\cite{mdm}, HOI-Diff~\cite{hoi-diff}, and ROG~\cite{rog}. HOI-Diff and ROG are consistent with our task setting. For MDM, we extended its output dimensions to obtain MDM*, enabling it to simultaneously predict human motion and object motion to meet the requirements of our task. Table \ref{tab:t2hoi} presents the quantitative comparison results on the two datasets. On the FullBodyManipulation dataset, our method outperforms the three baseline methods across all evaluation metrics. Compared to MDM* and HOI-Diff, our method achieves notable improvements, which can also be observed in the visualization results shown in Figure \ref{fig:t2hoi}. Compared with ROG, a state-of-the-art method specifically designed for this task, our method also achieves highly competitive performance. This strongly demonstrates the significant potential of our general framework when applied to a single task.


\begin{table*}[htbp]
    \centering
    \caption{Quantitative experimental results for the task of human motion generation guided by object motion(optionally with text) on the FullBodyManipulation dataset. }
    \begin{tabular}{c|lccccccc}
        \toprule
        & \textbf{Methods} & \textbf{HandJPE ↓} & \textbf{MPJPE ↓}
        & \textbf{\(C_{prec}\) ↑}  & \textbf{\(C_{rec}\) ↑}  & \textbf{\(C_{acc}\) ↑} & \textbf{ \(c \%\) ↑}  \\
        \midrule
        \multirow{3}{*}{w/o text} & OMOMO~\cite{omomo}      & 33.18 & 18.06 & 0.77 & 0.71 & 0.74  & 0.61 \\
        & SemGeoMo~\cite{semgeomo}      & 30.35 & 17.98 & 0.82 & 0.74 & 0.82   & 0.66 \\
        & Uni-HOI      & \textbf{29.87} & \textbf{17.63} & \textbf{0.85} & \textbf{0.75} & \textbf{0.84} & \textbf{0.70} \\
        \midrule
        \multirow{2}{*}{w text} & SemGeoMo-T~\cite{semgeomo}      & 27.84 & 16.62 & 0.84 & 0.74 & \textbf{0.85}  & 0.66\\
        & Uni-HOI & \textbf{27.56} & \textbf{16.41} & \textbf{0.86} & \textbf{0.76} & 0.84  &  \textbf{0.72}   \\
        \bottomrule
    \end{tabular}

    \label{tab:o2h}
\end{table*}

\begin{figure*}[htbp]
    \centering
    \includegraphics[width=0.77\textwidth]{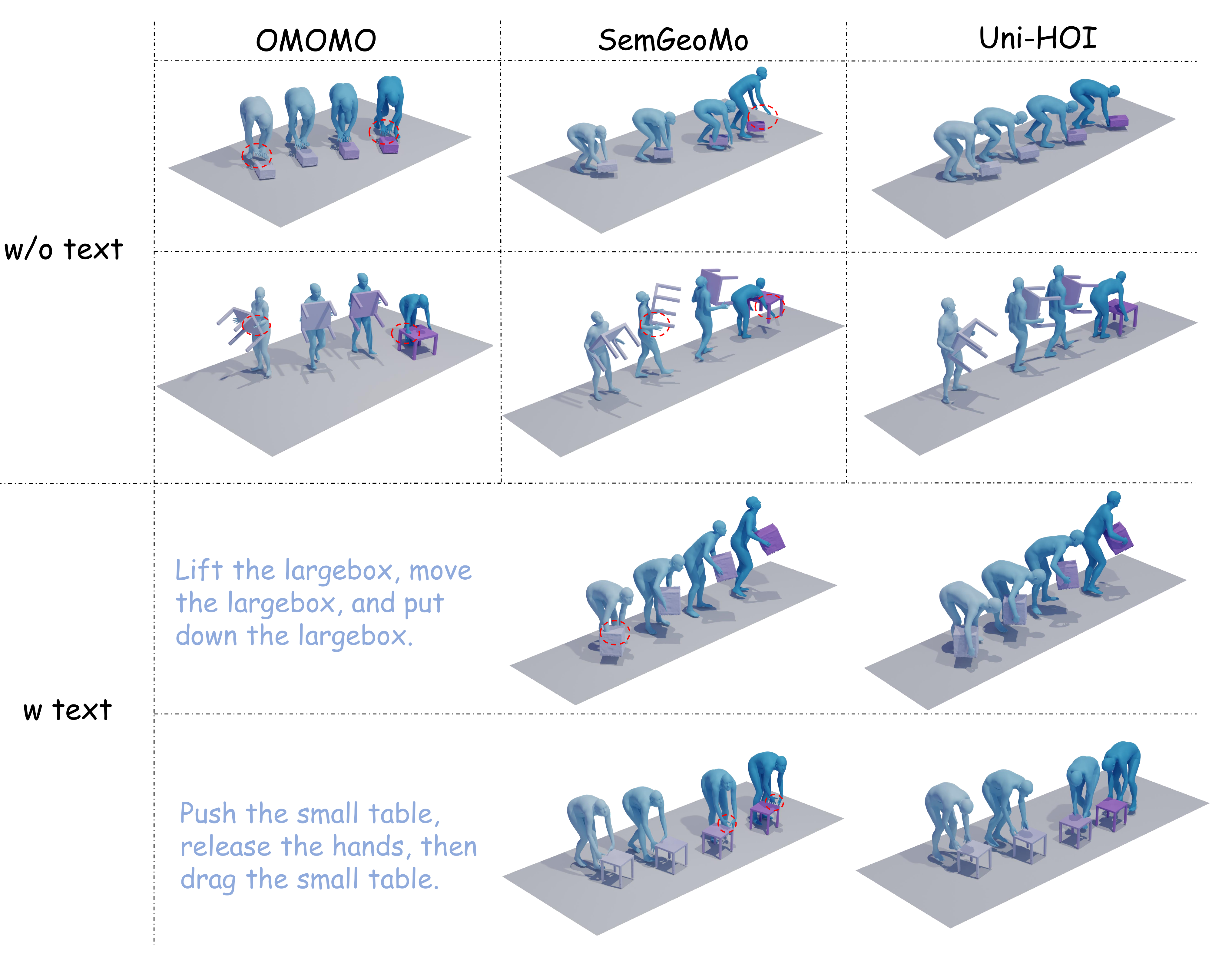}
    \caption{Visualization of comparative results for the task of human motion generation guided by object motion(optionally with text) on the FullBodyManipulation dataset.}
    \label{fig:o2h}
\end{figure*}



\begin{table}[htbp]
\centering
\caption{Quantitative comparison for human motion-driven object motion prediction on GRAB and BEHAVE datasets.}
\begin{tabular}{lcccc}
\toprule
\multirow{2}{*}{Method} & \multicolumn{2}{c}{GRAB} & \multicolumn{2}{c}{BEHAVE}  \\
\cmidrule(lr){2-3} \cmidrule(lr){4-5}
& Ec$\downarrow$ & Ev2v$\downarrow$ & Ec$\downarrow$ & Ev2v$\downarrow$  \\
\midrule
ObjPOP~\cite{objpop} & 0.047 & 0.244 & 0.147 & 0.472  \\
Uni-HOI & \textbf{0.024} & \textbf{0.126} & \textbf{0.092} & \textbf{0.325}  \\
\bottomrule
\end{tabular}
\label{tab:h2o}
\end{table}

\begin{figure*}[htbp]
    \centering
    \includegraphics[width=0.73\textwidth]{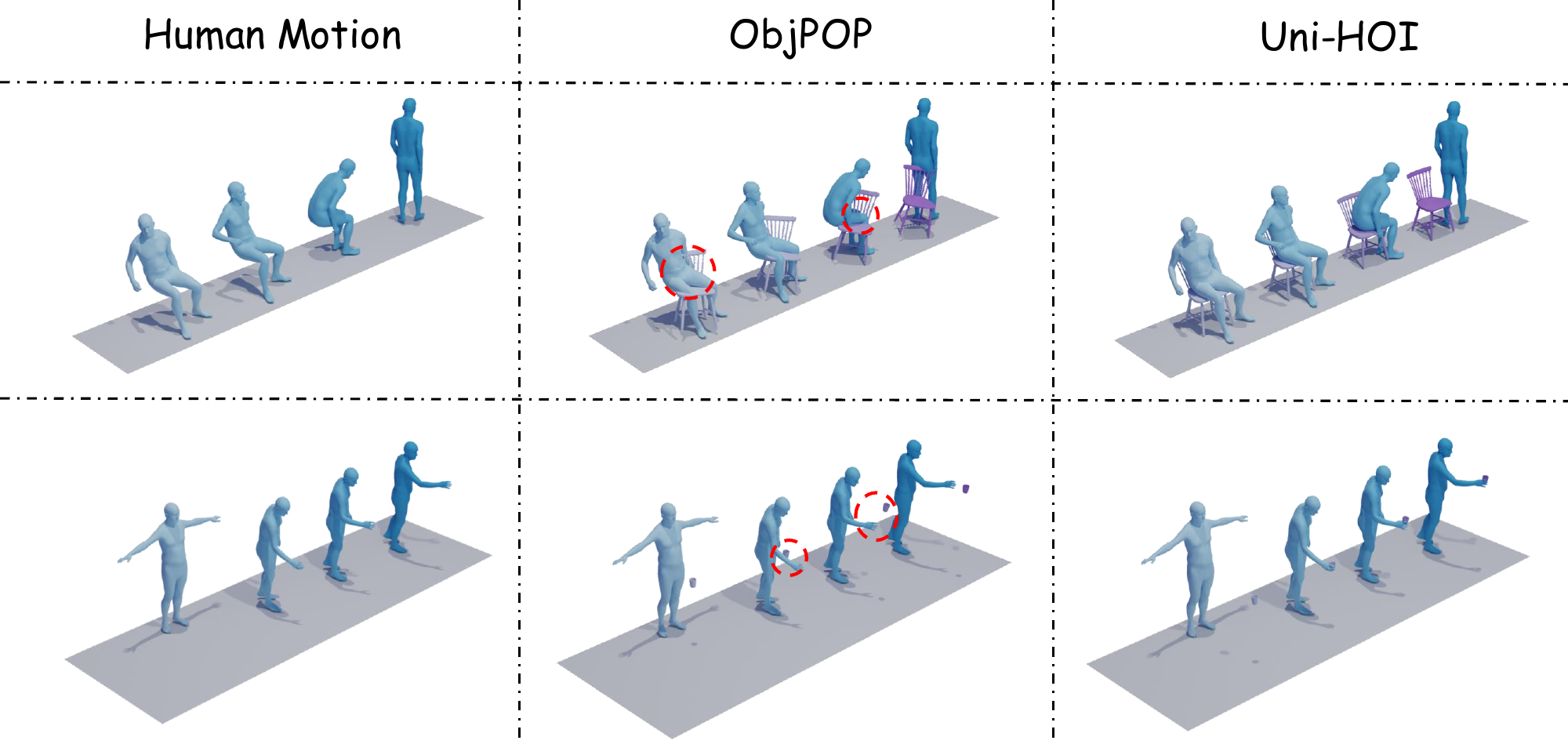}
    \caption{Visualization of comparative results for human motion-driven object motion prediction on the BEHAVE and GRAB datasets.}
    \label{fig:h2o}
\end{figure*}


\noindent \textbf{Comparisons on Object-to-Human.} For the task of human motion generation guided by object motion, OMOMO~\cite{omomo} was the first to explore this task and SemGeoMo~\cite{semgeomo} built upon it by introducing textual descriptions of human-object interaction. We also fine-tuned our model on the FullBodyManipulation dataset for this task and compared it with these two methods. Table \ref{tab:o2h} presents the corresponding quantitative results.  The experimental results show that our method achieves better performance than the two baseline methods across multiple evaluation metrics on this task. Furthermore, SemGeoMo can take ground-truth textual descriptions as input, using both object motion and text descriptions to predict human motion. Given this setting and the generality of our framework, we also conducted experiments on this task accordingly. As shown in the last two rows of Table \ref{tab:o2h}, our method still achieves superior performance on most metrics compared to SemGeoMo-T which denotes the variant using ground-truth text as input. Figure \ref{fig:o2h} presents the visual comparison results on both tasks, demonstrating that our method produces more natural and realistic results than SemGeoMo and OMOMO.

\noindent \textbf{Comparisons on Human-to-Object.} The task of predicting object pose from human interaction poses was first introduced by ObjPOP~\cite{objpop}, aiming to synthesize plausible 3D HOIs by predicting reasonable object poses. We explore a more general setting: predicting object motion from sequential human interaction motion. Using human motion as condition and object motion as output, we fine-tune our model on both GRAB~\cite{grab} and BEHAVE~\cite{behave} datasets. As a baseline, we adapt ObjPOP~\cite{objpop} to predict the object pose corresponding to each frame of human pose. Table \ref{tab:h2o} presents the quantitative comparison results on this task. Our method significantly outperforms the baseline on both datasets. Figure \ref{fig:h2o} shows the visualized comparison results between our model and baseline methods.

\subsection{Ablation Studies}
We conduct ablation experiments to investigate the impact of model size and training strategy on the final performance (detailed quantitative results can be found in the supplementary material). The experimental results show that, compared to using Qwen3-1B and Qwen3-4B, employing Qwen3-8B as the large language model backbone in our framework achieves better results across multiple evaluation metrics. In addition, compared to performing only the first-stage training on multiple tasks, second-stage fine-tuning on a single task significantly improves the performance of our framework on that specific task.

\begin{figure}[tbp]
    \centering
    \includegraphics[width=\columnwidth]{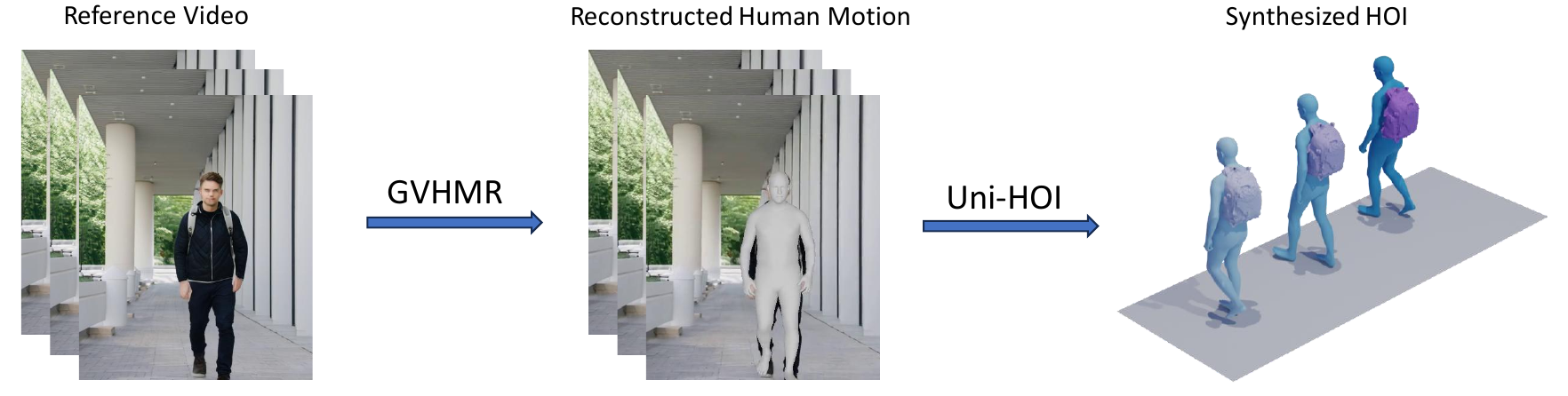}
    \caption{4D HOI synthesis. Uni-HOI can predict reasonable object motion for human in the reference videos.}
    \label{fig:application}
\end{figure}

\section{Applications}
Our model can be employed to reconstruct 4D Human-Object Interactions (HOI) from videos. Given a reference video, we first reconstruct the human pose parameters for each frame using GVHMR~\cite{gvhmr} to obtain human motion. Our framework then predicts a sequence of interactive object poses conditioned on the human motion (U(O|H)), thereby indirectly reconstructing complete 4D HOI from the video. 
The reference video can be obtained either from existing resources or generated using existing text-to-video techniques. In the latter case, our model can serve some zero-shot HOI generation scenarios.

\section{Limitations}
We have not extensively explored the combinatorial capabilities of more modalities, such as jointly predicting object poses conditioned on both human motion and text, or generating corresponding interaction captions from input human motion and object motion. Investigating these settings more comprehensively is an important direction for future work. Moreover, although our method already outperforms state-of-the-art (SOTA) approaches on several tasks without explicit human-object contact embeddings, incorporating explicit contact modeling could further improve performance.

\section{Conclusion}

In this work, we propose Uni-HOI, a unified framework that learns the joint distribution among text, human motion, and object motion in HOI. Unlike prior task-specific approaches, Uni-HOI leverages LLMs to establish a shared representation space across all three modalities. Two motion-specific VQ-VAEs convert heterogeneous motion data into token sequences compatible with LLM inputs, enabling a single model to handle arbitrary conditional inputs and outputs.
We introduce a two-stage training strategy: multi-task learning on a large-scale HOI dataset followed by task-specific fine-tuning. Extensive experiments on multiple tasks show that Uni-HOI consistently outperforms or matches task-specialized baselines using a single unified framework, demonstrating its effectiveness and generalization capability.
\bibliographystyle{ACM-Reference-Format}
\bibliography{sample-base}

\end{document}